# Solving Traveling Salesman Problem by Marker Method

**Masoumeh Vali**

Department of Mathematics, Dolatabad Branch, Islamic Azad University, Isfahan, Iran
E-mail: vali.masoumeh@gmail.com

**Abstract.** In this paper we use marker method and propose a new mutation operator that selects the nearest neighbor among all near neighbors solving Traveling Salesman Problem.

**Keywords:** Traveling Salesman Problem; marker method; mutation operator; adjacency matrix.

## I Introduction

The idea of the traveling salesman problem (TSP) is to find a tour of a given number of cities, visiting each city exactly once and returning to the starting city where the length of this tour is minimized.
The first instance of the traveling salesman problem was from Euler in 1759 whose problem was to move a knight to every position on a chess board exactly once ([8]).
The traveling salesman first gained fame in a book written by German salesman BF Voigt in 1832 on how to be a successful traveling salesman ([8]). He mentions the TSP, although not by that name, by suggesting that to cover as many locations as possible without visiting any location twice is the most important aspect of the scheduling of a tour. The origins of the TSP in mathematics are not really known all we know for certain is that it happened around 1931.
 The standard or symmetric traveling salesman problem can be stated mathematically as follows: Given a weighted graph G = (V, E) where the weight $c_{ij}$ on the edge between nodes $i$ and $j$ is a non-negative value, find the tour of all nodes that has the minimum total cost. The formulation of TSP is as follows [11]:

Minimize
$$\sum_{i<j} c_{ij} x_{ij} \qquad (1)$$

Subject to
$$\sum_{i<k} x_{ik} + \sum_{j>k} x_{kj} = 2 \quad (k \in V) \qquad (2)$$

$$\sum_{i,j \in S} x_{ij} \leq |S| - 1 \quad (S \subset V, 3 \leq |S| \leq n-3) \qquad (3)$$

$$x_{ij} = 0 \text{ or } 1 \quad (i,j) \in E \qquad (4)$$

In this formulation, constraints (2), (3) and (4) are referred to as degree constraints, subtour elimination constraints and integrality constraints, respectively. In the presence of (2), constraints (3) are algebraically equivalent to the connectivity constraints

$$\sum_{i\in S, j\in V\setminus S, j\in S} x_{ij} \geq 2 \quad (S \subset V, 3 \leq |S| \leq n-3)$$

**II Encoding**

For a TSP in which there are n cities we make a n× $n$ adjacency matrix which consist of a one in i, j th position if there is an arc from node i to node j that shows i, j cities are neighbor and have the smallest distance and a zero otherwise. This makes the initial solution for the problem. For example for 8 cities shown in Figure 1, city 1 is near to 8, 5 and 2, so in the matrix $a_{12} = a_{15} = a_{18} = 1$ and $a_{1j} = 0$ where $j \neq 2, 5, 8$. ( Figure 2)

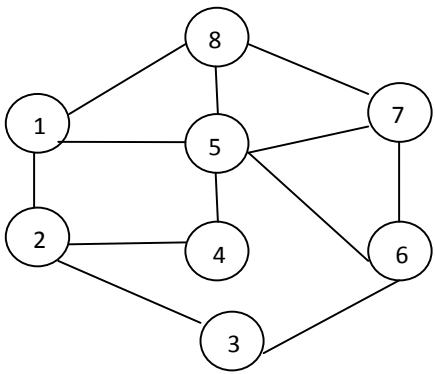

$$\begin{bmatrix} 0 & 1 & 0 & 0 & 1 & 0 & 0 & 1 \\ 1 & 0 & 1 & 1 & 0 & 0 & 0 & 0 \\ 0 & 1 & 0 & 1 & 0 & 1 & 0 & 0 \\ 0 & 1 & 1 & 0 & 1 & 0 & 0 & 0 \\ 1 & 0 & 0 & 1 & 0 & 1 & 1 & 1 \\ 0 & 0 & 1 & 0 & 1 & 0 & 1 & 0 \\ 0 & 0 & 0 & 0 & 1 & 1 & 0 & 1 \\ 1 & 0 & 0 & 0 & 1 & 0 & 1 & 0 \end{bmatrix}$$

Figure 1: TSP graph

Figure 2: Initial matrix

**III Mutation Operator**

In this method we start from first city located in first row of the matrix and check the smallest distance among elements that have 1 code then keep it and change others to zero. For instance, city 2 has the smallest distance to city 1 so for continuing the procedure, the algorithm goes to row to and evaluate the nearest city to it. Then it devotes 1 code to the answer and inverts others zero. After that it continues the method from city 4. Finally we get a matrix in which there is a 1 number in either row or column.
The solution shows a tour that arrangement of cities produces the best solution (lowest cost).
In the example, (1, 2, 4, 3, 6, 7, 8, 5) is the solution. (Figure 3)
Therefore, the matrix represents the tour that goes from city 1 to city 2, city 2 to city 4 and city 4 to city 3 and so forth.

$$\begin{bmatrix} 0 & 1 & 0 & 0 & 0 & 0 & 0 & 0 \\ 0 & 0 & 0 & 1 & 0 & 0 & 0 & 0 \\ 0 & 0 & 0 & 0 & 0 & 1 & 0 & 0 \\ 0 & 0 & 1 & 0 & 0 & 0 & 0 & 0 \\ 1 & 0 & 0 & 0 & 0 & 0 & 0 & 0 \\ 0 & 0 & 0 & 0 & 0 & 1 & 0 \\ 0 & 0 & 0 & 0 & 0 & 0 & 0 & 1 \\ 0 & 0 & 0 & 0 & 1 & 0 & 0 & 0 \end{bmatrix}$$

Figure 3: Final answer

## IV Conclusion

This method makes a Hamiltonian tour simply by devoting the number of each node (city) as a marker method. The matrix of distances is given as input information so D∗A gives the total cost of final solution where A is final answer and D is cost matrix.